\documentclass{article}

\usepackage[numbers]{natbib}
\usepackage{subfigure}
\usepackage{amsmath}
\usepackage{multirow}
\usepackage{amsfonts}
\usepackage{booktabs}

\usepackage[preprint]{neurips_2023}

\usepackage[utf8]{inputenc} 
\usepackage[T1]{fontenc}    
\usepackage{url}            
\usepackage{booktabs}       
\usepackage{amsfonts}       
\usepackage{nicefrac}       
\usepackage{microtype}      
\usepackage{xcolor}         
\usepackage{hyperref}       
\usepackage{graphicx}

\title{Beyond Surface Statistics:\\Scene Representations in a Latent Diffusion Model}

\author{%
  Yida Chen \\ 
  Harvard University \\ 
  Cambridge, MA 02138 \\
  \texttt{yidachen@g.harvard.edu}\\ 
  \And Fernanda Viégas\\
  Harvard University\\
  Cambridge, MA 02138 \\
  \texttt{fernanda@g.harvard.edu} 
  \And Martin Wattenberg \\
  Harvard University\\
  Cambridge, MA 02138 \\
  \texttt{wattenberg@g.harvard.edu} \\
}

\begin{document}

\maketitle

\begin{abstract}
  Latent diffusion models (LDMs) exhibit an impressive ability to produce realistic images, yet the inner workings of these models remain mysterious. Even when trained purely on images without explicit depth information, they typically output coherent pictures of 3D scenes. In this work, we investigate a basic interpretability question: does an LDM create and use an internal representation of simple scene geometry? Using linear probes, we find evidence that the internal activations of the LDM encode linear representations of both 3D depth data and a salient-object / background distinction. These representations appear surprisingly early in the denoising process---well before a human can easily make sense of the noisy images. Intervention experiments further indicate these representations play a causal role in image synthesis, and may be used for simple high-level editing of an LDM's output. 
\end{abstract}

\section{Introduction}

Latent diffusion models, or LDMs~\cite{rombach2022high}, are capable of synthesizing high-quality images given just a snippet of descriptive text. Yet it remains a mystery how these networks transform, say, the phrase “car in the street” into a picture of an automobile on a road. Do they simply memorize superficial correlations between pixel values and words? Or are they learning something deeper, such as an underlying model of objects such as cars, roads, and how they are typically positioned? 

In this work we investigate whether a specific LDM goes beyond surface statistics—literally and figuratively. We ask whether an LDM creates an internal 3D representation of the objects it portrays in two dimensions. To answer this question, we apply the methodology of linear probing~\cite{alain2016understanding} to a pretrained LDM~\cite{diffuser}. Our probes find linear representations of both a continuous depth map and a salient-object / background distinction. Intervention experiments further revealed the causal roles of these two representations in the model's output.

Our work fits into a long line of interpretability research. The general question of whether generative neural networks can be characterized as merely aggregating surface statistics is an area of heated debate in natural language processing~\cite{bender2020climbing,bisk2020experience}, with some evidence suggesting they do build world models~\cite{forbes2019neural,petroni2019language,li2022emergent}.

Investigations of image synthesis systems are less numerous, but suggest internal representations of scene geometry may play a role. For example, recent works leveraged pretrained diffusion models as priors when optimizing a neural radiance field network for 3D generations~\cite{poole2022dreamfusion, wang2022score}. Independent of our study, as part of an effort to steer denoising diffusion probabilistic models (DDPMs)~\cite{ho2020denoising}, Kim et al.~\cite{kim2022dag} found evidence of a complex non-linear representation of depth in a DDPM. Section~\ref{sec:related-work} discusses this and other related work in more detail.

\begin{figure}[t]
\centering
\includegraphics[width=0.55\textwidth]{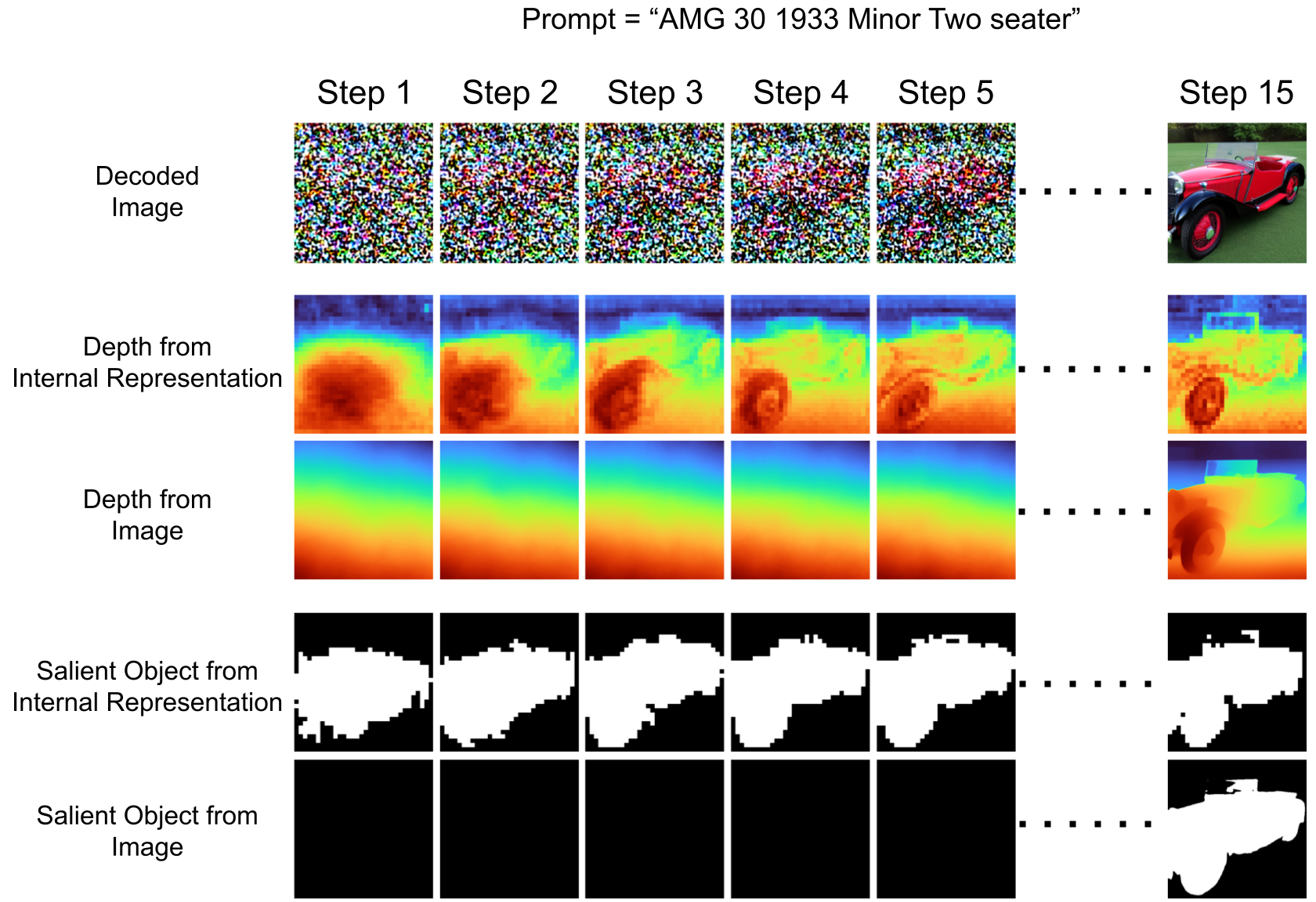}
\caption{LDM has an internal representation of depth and saliency that develops in the early denoising steps. Row 1: decoded images. Rows 2 \& 4 are predictions of probing classifiers based on the LDM's internal activations. Rows 3 \& 5 are baseline predictions from off-the-shelf models that take decoded RGB images decoded at each step as input. (See Appendix~\ref{appendix:vis-of-emerging-depth} for more examples.)}
\label{fig:overview-of-emerging-depth}
\vskip -0.05in
\end{figure}

Our findings---that a simple linear representation of 3D geometry plays a causal role in LDM image synthesis---provide additional evidence that neural networks learn to create world models in the context of image synthesis, and offer a basis for further study of the issue. 

\section{Background}
\textbf{Stable Diffusion:} We chose Stable Diffusion because it is an open-source text-to-image diffusion model used by millions of people~\cite{Fatunde.2022}. All of our experiments were conducted on the Stable Diffusion v1\footnote{Stable Diffusion v1 model: \href{https://github.com/CompVis/stable-diffusion}{github.com/CompVis/stable-diffusion}} that was trained without explicit depth information. Note that we did \textbf{not} use the subsequent version, which incorporated depth as a prior. 
 
Stable Diffusion is a two-stage image generative framework that consists of (1) a LDM $\epsilon_{\theta}$ for noise prediction and (2) a variational autoencoder (VAE) for converting data between latent and image space. It learned to generate images from noisy observations by reversing a forward diffusion process.

In the forward process, the encoder of VAE compresses a RGB image $x \in \mathbb{R}^{H \times W \times 3}$ to a latent variable $z \in \mathbb{R}^{\frac{H}{m} \times \frac{W}{m} \times 4}$ that resides in a low dimensional space. The forward diffusion transforms $z$ to a normally distributed variable via Gaussian transition.  

The LDM $\epsilon_{\theta}$ is trained to predict the noise added during the forward diffusion. In the generative process, $\epsilon_{\theta}$ synthesizes a $z$ by progressively denoising a sampled noisy observation $z_{T} \sim \mathcal{N}(0, \text{I})$. The decoder of VAE converts the denoised latent $z$ to the image space.

To allow conditional sampling $p(z|y)$, the LDM fuses contextual embeddings $\tau(y)$ through its cross-attention layer~\cite{vaswani2017attention}. Our experiments study the text-to-image variant of Stable Diffusion, where $y$ is a prompt in a word embedding space and $\tau$ is the CLIP text encoder~\cite{radford2021learning}. 

\textbf{Probing Classifiers} are used to interpret the representation learned by deep neural networks~\cite{alain2016understanding, belinkov2021probing}. We borrow this technique to investigate whether an LDM builds a representation of per-pixel depth. 

Probing classifiers take the intermediate activation of a neural network as input to infer some properties. We define the intermediate activation as the model's internal representation. A high prediction accuracy indicates a strong correlation between the learned representations of the model and the predicted property.

The probing classifier may find spurious correlations when the model has an enormous feature space. To quantify the strength of correlation found by the classifier $p$, previous works test $p$ on the internal representation of a randomized untrained model in addition to the trained version~\cite{conneau2018you,zhang2018language,chrupala2020analyzing}. The performance on the untrained model serves as a baseline for the probing task.

Nevertheless, high probing performance does not imply a causal relation between the representations of model and the property of its output. To uncover a causal link, recent works use the projection learned by $p$ to modify the internal activations of model and measure the effects on its outputs~\cite{tucker2021if,elazar2021amnesic}. We adopt the approach above in designing our intervention experiments~\ref{sec:intervention-binary}.

\section{Probing for discrete and continuous depth}
Stable Diffusion often creates scenes that appear to have a consistent 3D depth dimension, with regions arranged from closest to farthest relative to a viewpoint. However, besides this continuous depth dimension, we also see images with Bokeh effects, where some objects are in clear focus and their background surroundings are blurred. We therefore explored the world representations of depth from two perspectives: (1) a \textbf{discrete binary depth} representation from the perspective of human cognition, where each pixel either belongs to certain visually attractive objects or their background, and (2) a \textbf{continuous depth} representation from the perspective of 3D geometry, where all pixels have a relative distance to a single viewpoint.

We formulated probing for binary depth as a salient object detection task~\cite{lee2022tracer}, which aims to segment the objects that are most distinctive to human vision. The probing for continuous depth dimension can be characterized by a monocular depth estimation task~\cite{lasinger2019towards}.

Recent works have shown that vision transformers outperformed convolutional neural networks in depth estimation tasks~\cite{johnston2020self,ranftl2021vision}. As part of our work on depth probing, we also pursued a preliminary study where we examined the depth representations in convolutional layers but found they were generally weaker than the those in self-attention layers (see Appendix~\ref{appendix:depth-in-conv-vs-sa}). Our study thus focused on the representation learned by global self-attention layers when exploring the depth information encoded in Stable Diffusion.  

\subsection{Synthetic dataset}
\label{sec:synthetic-dataset}
For our probing experiments, we created a dataset of synthesized images paired with their input latents $z_{T} \sim \mathcal{N}(0, 1)$, prompts $y$, and depth labels $d$. For this synthetic dataset, we generated 1000 images using a pretrainedStable Diffusion model\footnote{We initialized Stable Diffusion with official \href{https://huggingface.co/CompVis/stable-diffusion-v1-4}{checkpoint v1.4}.} and linear multi-step scheduler~\cite{karras2022elucidating} with 15 steps.

To ensure the quality and diversity of synthesized images, we sampled prompts from a partition of the LAION-AESTHETICS v2 dataset~\cite{schuhmann2021laion}, the same dataset used for fine-tuning the pretrained Stable Diffusion model. Each image is generated using a different prompt and latent input.

Since the output images do not have ground truth labels for salient-object / background, we instead synthesized them using an off-the-shelf salient object tracing model TRACER~\cite{lee2022tracer}. Similarly for depth, we estimated the relative inverse depth map\footnote{Depth output by MiDaS is proportional to the inverse of absolute distance $\frac{1}{D}$ between the view and a region.} of output images using the pretrained MiDaS model~\cite{lasinger2019towards}. To our knowledge, MiDaS and Tracer are the best off-the-shelf models for monocular depth estimation and salient object detection, respectively.

In creating this synthetic dataset, we had to address problematic images. These ranged from offensive content to images with corrupted or distorted objects. For ethical concerns, we manually scrutinized the synthesized dataset and discarded images with explicit sexual and violent content. It is also challenging to infer depth from corrupted images using external models trained on natural images. For controlling quality of synthesized labels, we excluded images that were visually corrupted ($\sim10\%$ of dataset). Finally, we took out images without the depth concept, such as black-and-white comic art. After generating the labels, we removed the images where TRACER failed to identify any salient object. The resulting dataset has 617 samples. We split it into a training set with 246 samples and a test set with 371 samples for the following experiments.

\subsection{Binary depth: salient object and background}
\label{sec:probe-binary-depth}
To investigate discrete binary depth representations inside the LDM, we extract the intermediate output $\epsilon_{\theta(l,t)} \in \mathbb{R}^{h_l \times w_l \times c}$ of its self-attention layer $l$ at sampling step $t$. A linear classifier $p_b$ is trained on $\epsilon_{\theta(l,t)}$ to predict the pixel-level logits $\hat{d_b} \in \mathbb{R}^{512 \times 512 \times 2}$ for salient and background classes. This probing classifier can be written as:
\begin{gather}
    \hat{d_b} = \text{Softmax}(\text{Interp} (W \epsilon_{\theta(l,t)}, \frac{512}{h_l}, \frac{512}{w_l}))
\end{gather}
The LDM's self-attention layers operate at different spatial and feature dimensions (see Appendix~\ref{appendix:spatial-feature-dimension-of-self-attention-layers}). We interpolate lower resolution predictions $W \epsilon_{\theta(l,t)}$ to the size of synthesized images. $W \in \mathbb{R}^{c \times 2}$ are the trainable parameters of probing classifier. $W$ is updated using gradient descent that minimizes the Cross Entropy loss between $\hat{d_b}$ and the synthetic label $d_b$. 

We measured the segmentation performance using the Dice coefficient between the predictions $\Tilde{d_b} = arg~max(\hat{d_b})$ and the synthetic ground truth $d_b$, $D(\Tilde{d_b}, d_b) = \frac{2(\Tilde{d_b}~ \cap~ d_b)}{| \Tilde{d_b} |~ +~| d_b |}$. 

\subsection{Continuous relative depth information}
\label{sec:probe-monocular-depth}
We are also interested if a more fine-grained depth dimension also exists inside LDM. Similar to the probing of binary depth, we extract the output from self-attention layers and train a linear regressor on them to predict the MiDaS relative depth map $d_c \in \mathbb{R}^{512 \times 512 \times 1}$.
\begin{gather}
    \hat{d_c} = \text{Interp} (W \epsilon_{\theta(l,t)}, \frac{512}{h_l}, \frac{512}{w_l})
\end{gather}
We update $W \in \mathbb{R}^{c \times 1}$ by minizing the Huber loss between $\hat{d_c}$ and $d_c$. We also experimented with regularizing depth predictions with a smoothness constraint~\cite{johnston2020self}. However, the smoothness constraint had a negative impact on probing (see Appendix~\ref{appendix:smoothness-regularization}). Following existing practices~\cite{eigen2014depth, lasinger2019towards}, the accuracy of depth probing is measured using root mean squared error.

We probed the internal representations of salient regions and depth across all self-attention layers $\{l_1, l_2, \ldots, l_{16} \}$ of LDM at all sampling steps ($t = \{ 1, 2, \ldots, 15\}$). Probing classifiers were trained separately on each layer and step. For a fair comparison, we fixed the number of training epochs, optimization algorithm and learning rate for all training.

\textbf{Controlled Probing Experiment:} Because of the large feature space of LDM, probing classifiers may find a spurious representation of depth. We need a baseline to quantify the strength of the internal representation found by probing classifiers. For both probing tasks, we measured the baseline performance using probing classifiers trained on the internal representations of a randomized LDM.

\section{Linear Representations of Depth and Salient Objects}

\begin{figure}[t]
\begin{center}
\includegraphics[width=0.90\columnwidth]{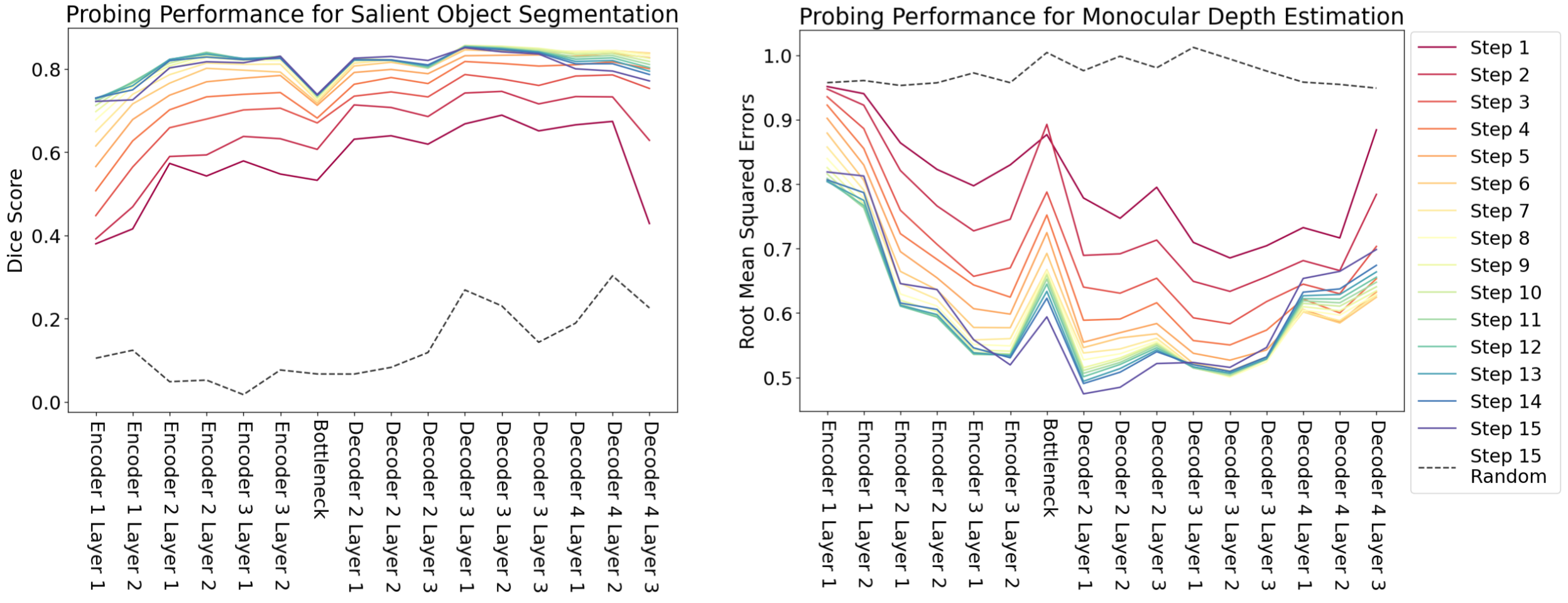}
\caption{Probing performance of salient object segmentation and depth estimation on test set. For both tasks, performance grew rapidly in the early denoising process and converged after step 5. The black dashed line is the baseline performance measured on a randomized LDM at step 15, significantly lower than its trained counterpart.
}
\label{fig:probing-performance}
\end{center}
\vskip -0.05in
\end{figure}

For both probing tasks, we obtained high probing performance using the internal representations of LDM, especially in the later denoising steps. As shown in Figure~\ref{fig:probing-performance}, the performance increased significantly in the first 5 steps and gradually converged in the late denoising process. At the last step, probing classifiers achieved an average Dice coefficient of $0.85$ for salient object segmentation and average RMSE of $0.47$ for depth estimation when inputted with decoder side self-attention.

\begin{figure}[t]
\begin{center}
\includegraphics[width=0.95\textwidth]{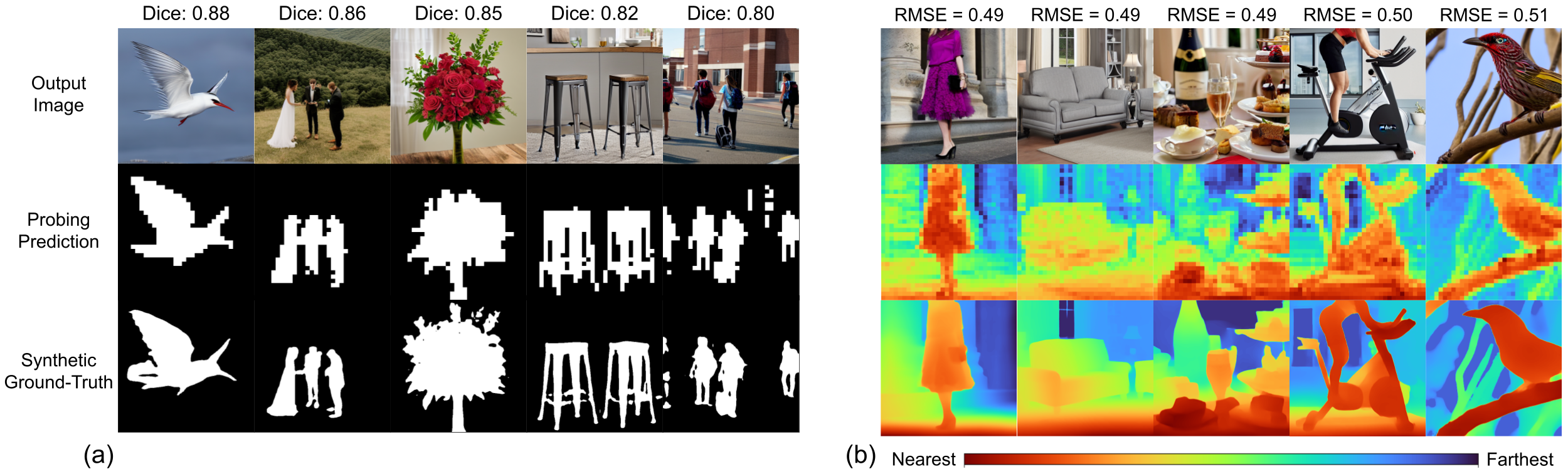}
\caption{Row 2 shows the predictions of two probing classifiers. The probing classifiers achieved an average Dice score of $0.85$ for salient object segmentation and an average RMSE of $0.50$ for depth estimation on 371 test samples.}
\label{fig:foreground-segmentation-masks-depth-estimations}
\end{center}
\vskip -0.1in
\end{figure}

The Dice coefficient suggests that our probing classifiers achieved high-quality salient object segmentation in the late denoising steps. As Figure~\ref{fig:foreground-segmentation-masks-depth-estimations}a shows, the probing classifier captured salient objects in various scenes when inferring on a decoder side self-attention output at the last denoising step. Even though the spatial dimension of input representations is only $\frac{1}{16} \times \frac{1}{16}$ of output images, the predicted masks still catch fine-grained details such as limbs of human and legs of chairs. Our synthetic dataset covers a variety of objects, including humans, vehicles, furniture, animals, machinery, etc. It's unlikely the linear classifiers learned to select features corresponding to certain kinds of objects or memorize their locations in images. Probing classifiers also obtained accurate depth estimations at late denoising steps. As shown in Figure~\ref{fig:foreground-segmentation-masks-depth-estimations}b, the probing predictions matched the synthetic depth maps for images of both indoor and outdoor scenes.

In the controlled experiment, the probing classifiers performed significantly worse when trained on the randomized LDM. At the last denoising step, the highest Dice coefficient achieved on the randomized LDM was only $0.30$ and the smallest RMSE in depth estimation was only $0.95$. These substantial differences indicate the internal representations of trained LDM have a strong correlation with the depth which cannot be easily captured by a random model.

\subsection{Depth emerges at early denoising steps}
\label{sec:depth-emerges-early}
We saw a dramatic increase in probing performance during the first 5 steps. For both probing tasks, the performance difference between successive denoising steps vanished after step 5. High probing performance at the early steps suggests an interesting behavior of LDM: the depth dimension develops at a stage when the decoded image still appears extremely noisy to a human.

\begin{figure}[t]
\vskip -0.15in
\begin{center}
\includegraphics[width=0.99\textwidth]{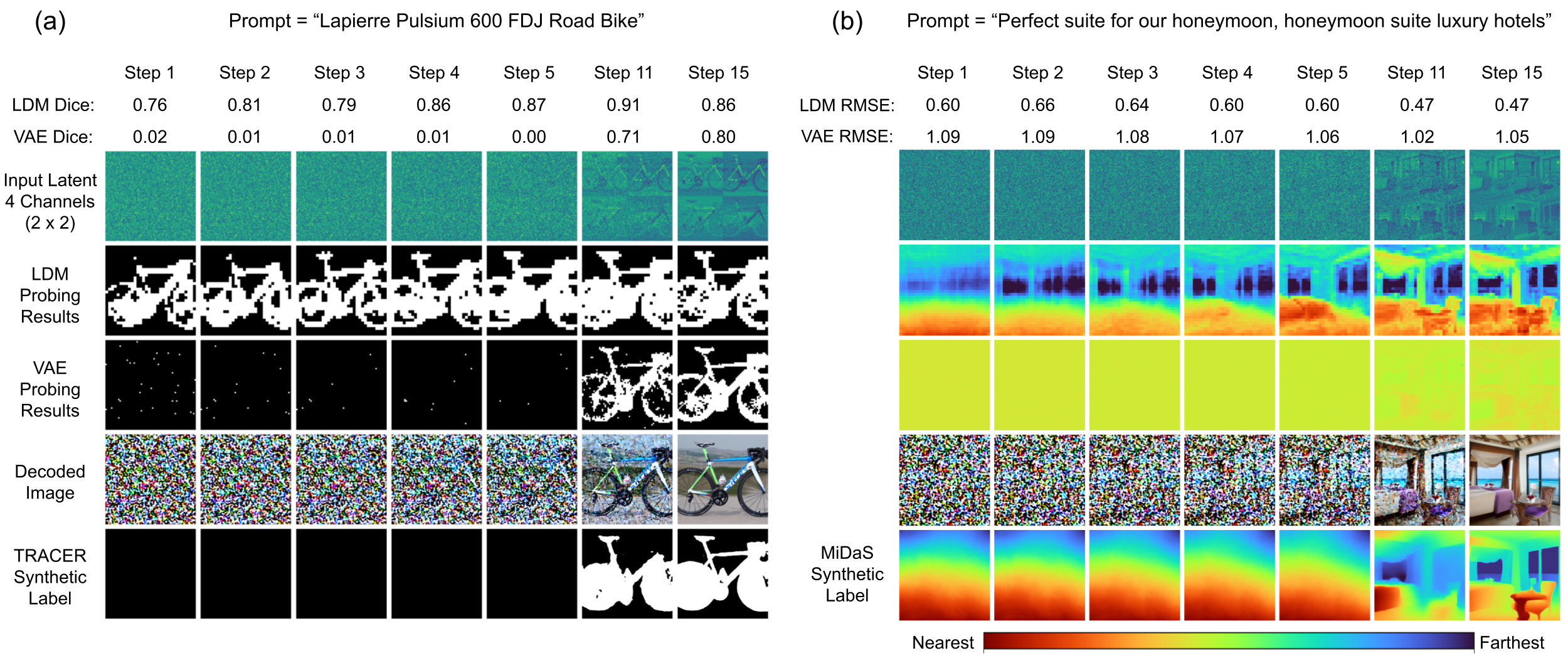}
\caption{LDM's internal representations of salient object (a) and depth (b) appear surprisingly early in the denoising. Probing the internal representations of VAE, however, cannot find the salient objects and depth when latents are corrupted. We reported the Dice coefficient and RMSE between probing results at each step and the synthetic label obtained at step 15.}
\label{fig:emerging-binary-depth-dimension}
\end{center}
\vskip -0.1in
\end{figure}

As Figure~\ref{fig:emerging-binary-depth-dimension}a shows, the images decoded at early denoising steps are heavily corrupted. A human viewer can barely capture any meaningful information from them. Unexpectedly, the representation of LDM delineates the salient object in the second denoising step. For comparison, we ran TRACER on the images decoded from partially denoised latents. This state-of-the-art salient object detection model cannot find the salient objects inside noisy images. 

A continuous depth dimension also appears at the earlier steps. Figure~\ref{fig:emerging-binary-depth-dimension}b shows that the internal representation has determined the layout of the hotel room as early as at step 5. We ran MiDaS on the partially denoised images, but it failed until a significant amount of noise was removed (at step 11).

\begin{table}[t]
\caption{Weak depth information was found in VAE's bottleneck activations, but LDM encodes a much stronger representation of depth. This table shows the probing performance achieved by the LDM's and VAE's self-attention layers at steps 5 and 15.}
\begin{center}
\begin{footnotesize}
\begin{tabular}{cccc}
\toprule
Step                & \multicolumn{1}{l}{Model} & \begin{tabular}[c]{@{}c@{}}Saliency Detection \\ (average Dice $\uparrow$)\end{tabular} & \begin{tabular}[c]{@{}c@{}}Depth Estimation \\ (average RMSE $\downarrow$)\end{tabular} \\ \midrule
\multirow{2}{*}{5}  & LDM                         & \textbf{0.84}                                                                & \textbf{0.53}                                                                 \\
                    & VAE                         & 0.15                                                                 & 0.96                                                                 \\ \midrule
\multirow{2}{*}{15} & LDM                         & \textbf{0.85}                                                                  & \textbf{0.47}                                                                 \\
                    & VAE                         & 0.71                                                                 & 0.84                                                                 \\ \midrule
\end{tabular}
\end{footnotesize}
\end{center}
\label{table:depth-representation-in-vae}
\end{table}

\subsection{Is depth represented in latent space?}
Because an LDM operates on a latent-space representation of an image, one might ask whether this representation itself contains depth information. To address this question, we performed a probing study of the VAE self-attention layer.

For the salient object detection task, the self-attention layer in VAE cannot decode salient objects from the corrupted latents at early steps (see Table~\ref{table:depth-representation-in-vae}). Its performance starts to improve when details in latent vectors become more perceptible. After the latent is completely denoised at step 15, the segmentation performance of VAE self-attention is slightly lower than the average performance of LDM decoder side self-attention (Dice coefficient of 0.71 vs 0.85). For the depth estimation, the self-attention layer of VAE failed across all steps. The average RMSE obtained by VAE self-attention at the final step is still 0.84.

These results suggest that the VAE bottleneck does not contain a significant linear representation of depth early in the denoising process. Later in the process, some saliency / background information emerges. In all cases, it seems the LDM has a stronger representation of depth.

\section{Causal Role of Depth Representation}
\label{sec:probe-causal-intervention}
Probing experiments show a high correlation between the internal representations of LDM and the depth of its output images. But does this representation play a causal role? To answer this question, we designed causal intervention experiments for salient (foreground) object and depth representations. Essentially, we want to change the model's output by solely modifying the internal representations in LDM using the projection learned by our probing classifiers. 

\begin{figure*}[t]
\begin{center}
\centerline{\includegraphics[width=0.9\textwidth]{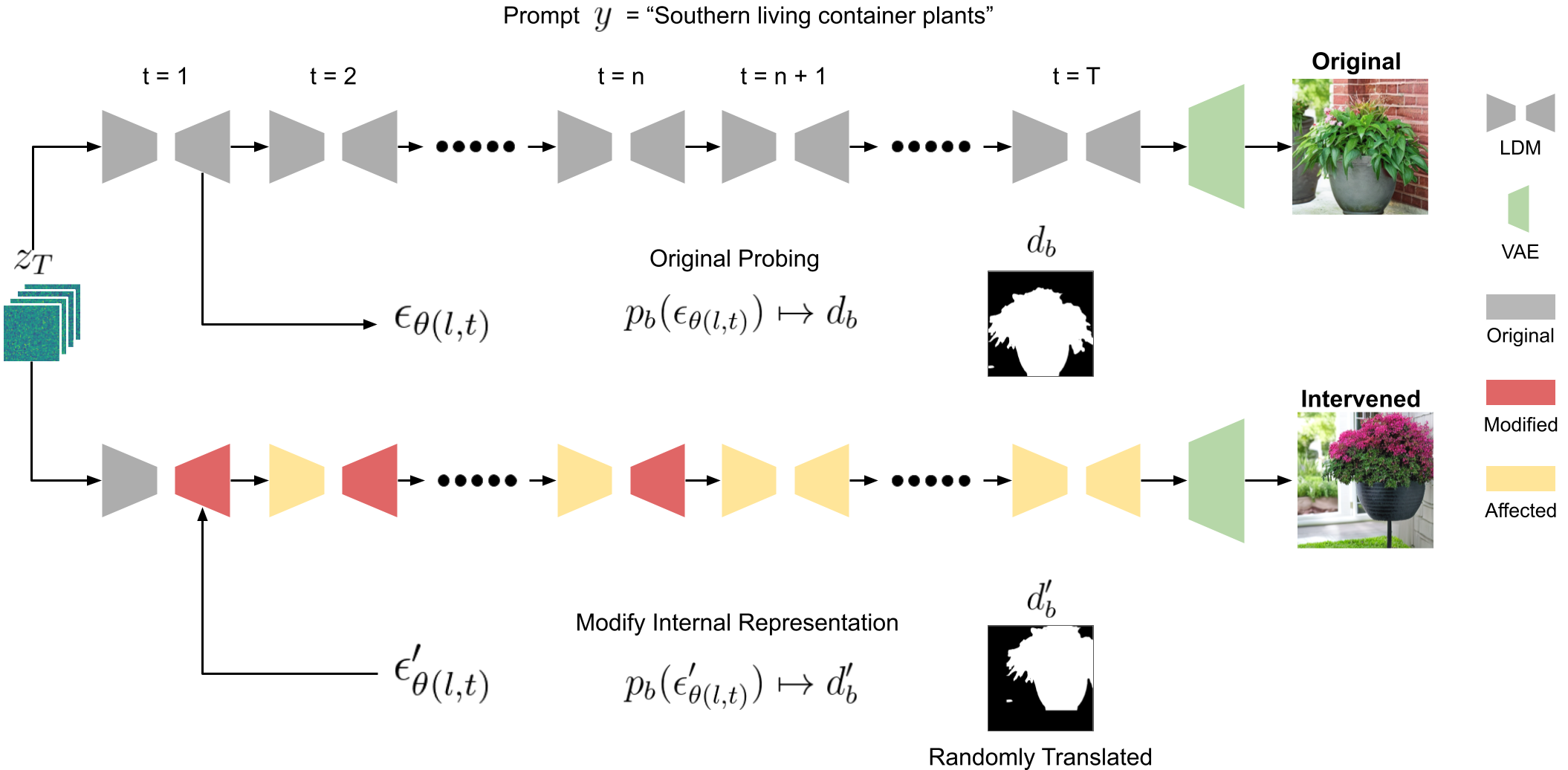}}
\caption{This figure illustrates our intervention workflow, where the foreground object was repositioned in the intervened output. When modifying the representations at a chosen layer (in red), we interrupt LDM before the layer forwards its activation $\epsilon_{\theta(l, t)}$. $\epsilon_{\theta(l, t)}$ is updated so the probing classifier's prediction changes to match the modified label. We replace $\epsilon_{\theta(l, t)}^{\prime}$ with the original activations and resume the denoising process. Since LDM uses the latents denoised from previous step as input, activations after the intervened layers are also affected by the modification (highlighted in yellow). We adopt a similar intervention scheme for the depth representation.}
\label{fig:intervention-example-workflow}
\end{center}
\vskip -0.2in
\end{figure*}

\begin{figure}[t]
\begin{center}
\centerline{\includegraphics[width=0.55\columnwidth]{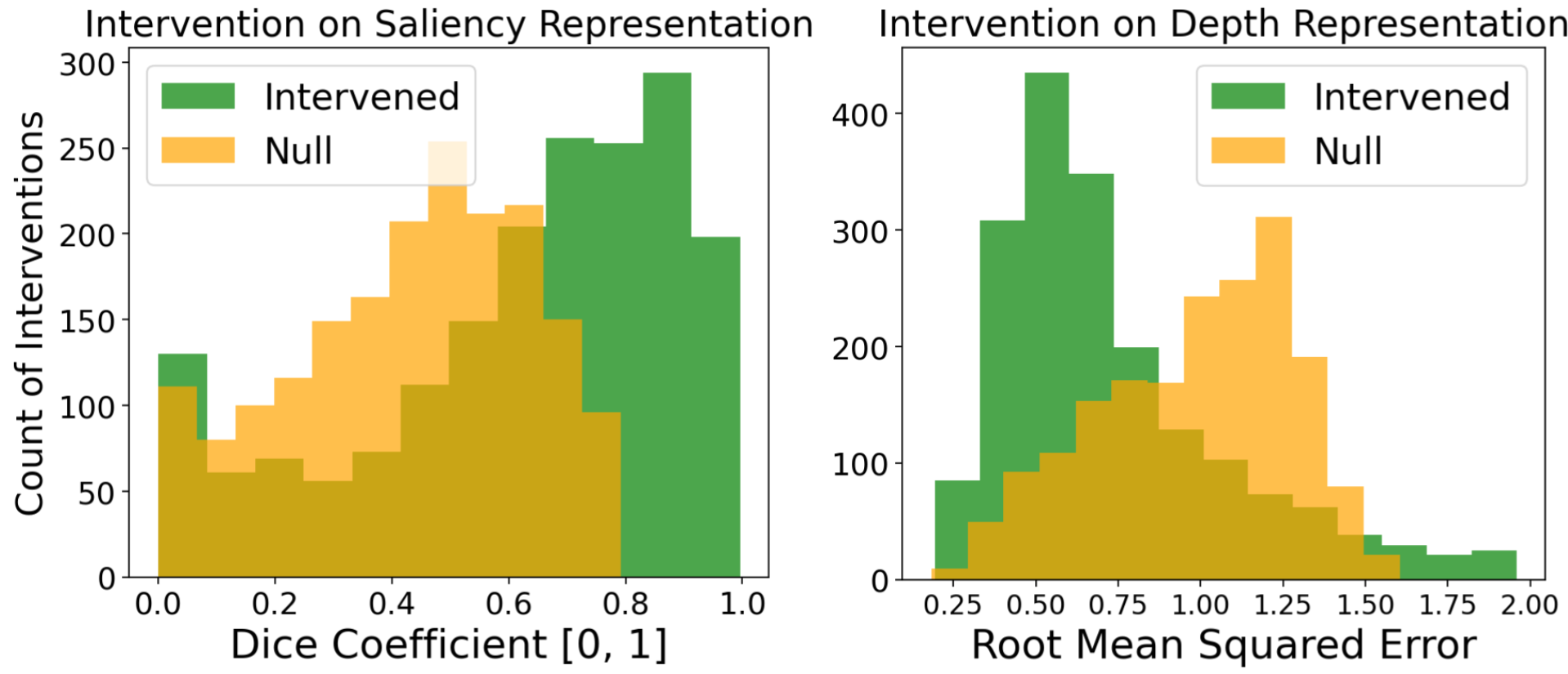}}
\vskip 0.1in
\begin{small}
\begin{tabular}{ccc}
Representations & \multicolumn{1}{l}{Pre-intervention} & \multicolumn{1}{l}{Post-intervention} \\ \hline
Saliency (Dice~$\uparrow$) & 0.46                                 & 0.69                                  \\
Depth (RMSE~$\downarrow$)  & 1.02                                 & 0.63                                  \\ \hline
\end{tabular}
\end{small}
\caption{Quantitative results show that intervening saliency representation had causal effects on model's outputs. The median Dice coefficient between modified salient object masks and synthetic masks of 1851 intervened outputs is 0.69 (vs. 0.46 of null baseline). The median RMSE between modified depth maps and depth maps of 1855 intervened outputs is 0.63 (vs. 1.02 of null baseline).}
\label{fig:intervention_stats}
\end{center}
\vskip -0.1in
\end{figure}

\subsection{Intervention experiment}
\label{sec:intervention-binary}
\textbf{Intervention: Repositioning foreground objects.} Our goal is to see if changing the depth representation, with the same prompt and initial input, will lead to a corresponding change in apparent depth in the output image. To modify the geometry while preserving the semantics of original output, we test a minimal change: we identify a foreground object, and translate its representation in 2D space. 

When translating the object's representation, we used a modified salient object mask $d_{b}^{\prime}$ as the reference. The mask $d_{b}^{\prime}$ was generated by randomly translating the original mask $d_b$ with vertical and horizontal translations sampled from $\mathcal{U}(-120, -90) \cup (90, 120)$. We created 5 different $d_{b}^{\prime}$ for each sample in the test set. 

The intervention then modifies the LDM's representation so the probing classifier's output, if using a modified representation as input, changes to $d_{b}^{\prime}$. This is achieved by updating the internal representation $\epsilon_{\theta(l,t)}$ using gradients from the probing classifier $p_b$. The weights of $p_b$ are frozen.
\begin{gather}
\label{eq:gradient-intervention}
    g = \frac{\partial \mathcal{L}_{CE}(p_b(\epsilon_{\theta(l,t)}), d_{b}^{\prime})}{\partial \epsilon_{\theta(l,t)}}
\end{gather}
$\epsilon_{\theta(l,t)}$ is updated using Adam~\cite{kingma2014adam} with gradients calculated by Eq.\ref{eq:gradient-intervention}.

In Section~\ref{sec:depth-emerges-early}, we observed that the representation of salient objects formed very early on. Interventions on the later denoising process cannot effectively change the position of foreground objects. The decoder side self-attention also performed better than the encoder side ones in the early steps. Thus, during intervention, we modified the activations of decoder side self-attention layers at the first five steps. In a preliminary study, we experimented with intervening in different number of denoising steps for saliency and depth representations (see Appendix~\ref{appendix:choice-of-intervened-denoising-steps}). For each intervened layer at the chosen denoising steps, we optimized $\epsilon_{\theta(l,t)}$ so the probing prediction $p_b(\epsilon_{\theta(l,t)}^{\prime})$ changed to match $d_{b}^{\prime}$. 

To assess the effects of this intervention, we computed the Dice coefficient between the modified mask and synthetic mask of intervened output. If the saliency representation has a causal role in LDM, the salient regions of newly generated images should match the modified masks.

We performed a similar intervention on the continuous depth representation. The depth map of the original output $d_c$ was translated with horizontal and vertical translations sampled from $\mathcal{U}(-120, -90)\cup(90, 120)$ to generate the modified map $d_{c}^{\prime}$. Empty areas outside translated depth map were filled with its edge values. As in the intervention for saliency, $\epsilon_{\theta(l,t)}$ is updated using the gradients of probing regressor $p_c$ so its output matches $d_{c}^{\prime}$. We calculated the gradients using the same Eq.~\ref{eq:gradient-intervention} with $\mathcal{L}_{CE}$, $p_b$, and $d_{b}^{\prime}$ replaced by $\mathcal{L}_{Huber}$, $p_c$, and $d_{c}^{\prime}$. We intervened on all self-attention layers at the first three sampling steps. The intervention on the depth representations was more effective when modifying all self-attention layer activations.

The effects of our interventions were measured by the RMSE between $d_{c}^{\prime}$ and the depth map of intervened output. If a causal role exists, the depth maps of new outputs should match the modified depth maps used in intervention.

\begin{figure}[t]
\begin{center}
\includegraphics[width=0.47\columnwidth]{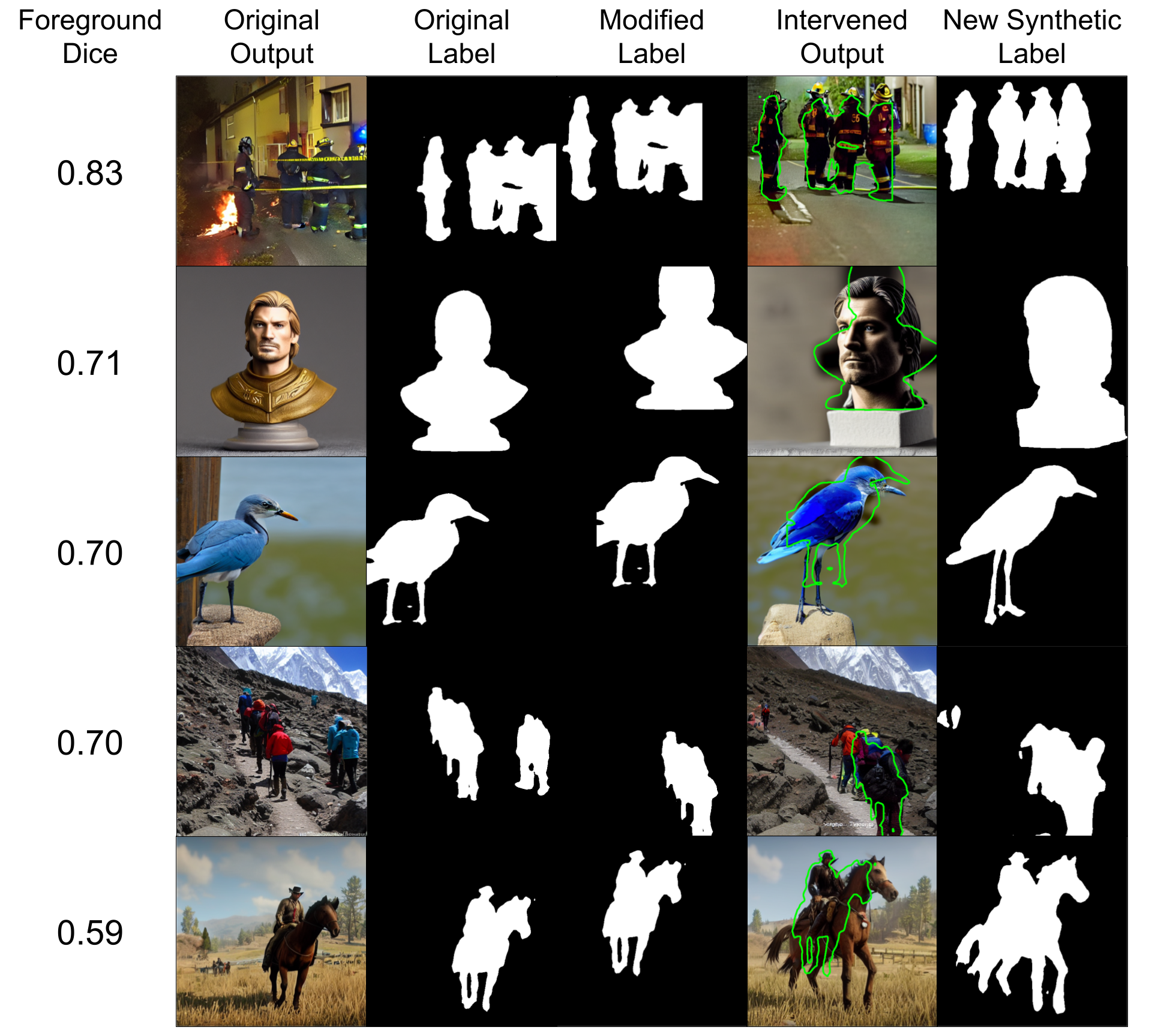}
\hspace{0.03\columnwidth}
\includegraphics[width=0.47\columnwidth]{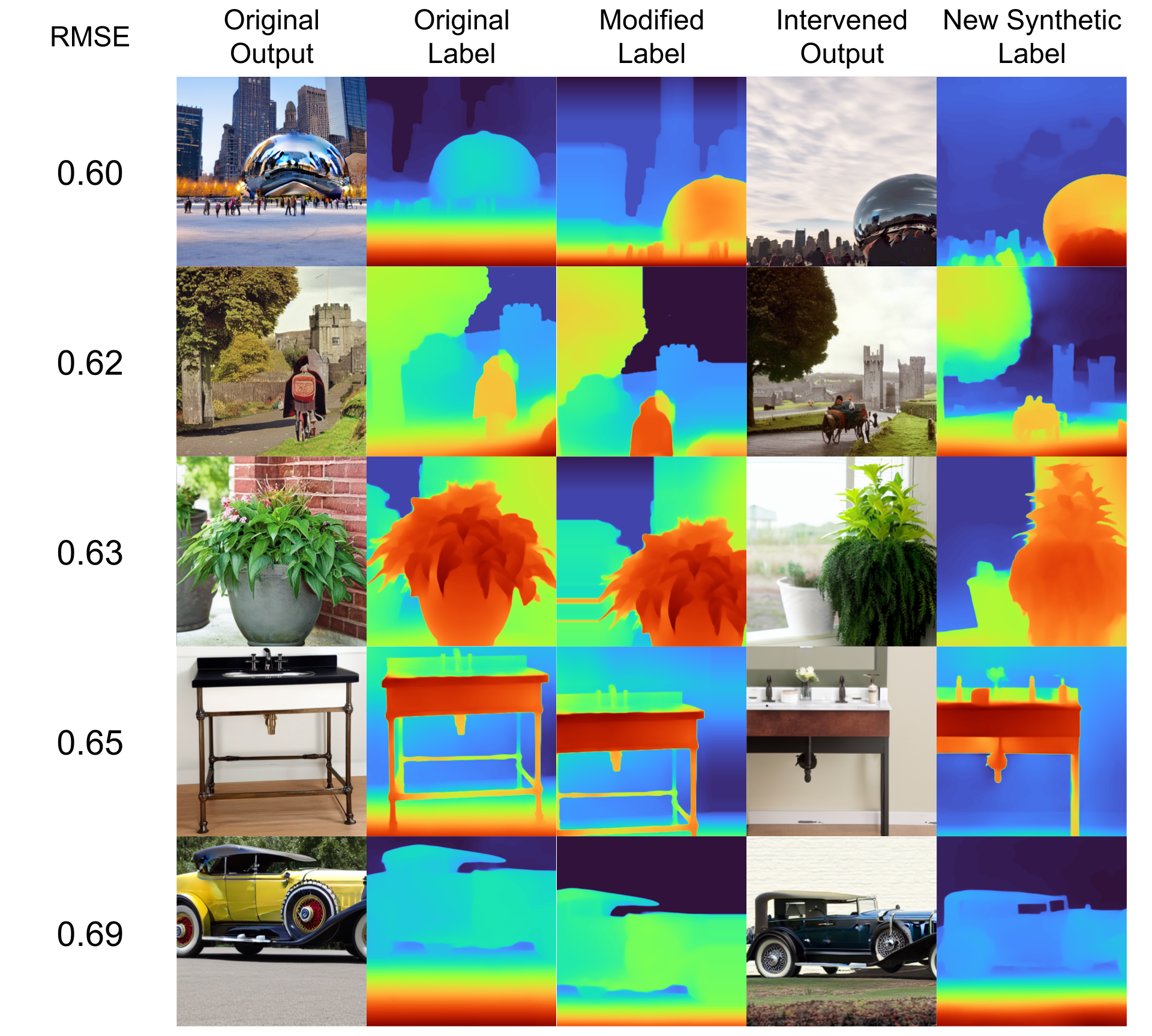}
\caption{ 
Column 1 \& 2: model's original outputs and labels of salient objects or depth; Column 3: the modified labels used in intervention; Column 4: the intervened outputs (with contours of modified labels in green for intervention on saliency); Column 5: synthetic labels of intervention outputs. Dice and RMSE are measured between modified labels and synthetic labels of intervention outputs.}
\label{fig:intervention_binary_examples}
\end{center}
\vskip -0.15in
\end{figure}

\textbf{Salient object intervention results:} We conducted our proposed intervention on the test set of our synthetic dataset (371 samples). For the resultant 1855 interventions, the median Dice coefficient is 0.69 between the modified salient object labels and the synthetic labels of intervention outputs. We further compared the modified label $d^{\prime}_b$ with the synthetic label $d_b$ of the original images, which acts as a baseline. The comparison gave us a median Dice overlap of 0.46 (see Figure~\ref{fig:intervention_stats}).

As Figure~\ref{fig:intervention_binary_examples} shows, our interventions successfully repositioned the foreground objects by solely modifying the self-attention activation using the projection learned by probing classifiers. Our results suggest the representation of salient objects has a causal role on the model's output.

\textbf{Depth intervention results:}
The median RMSE for 1855 intervention outputs is 0.63, whereas the median RMSE of the null intervention is 1.02 (between $d_{c}^{\prime}$ and the original depth $d_c$). This result confirmed the causal role of depth representations. In a fine-grained intervention experiment (see Appendix~\ref{appendix:fine-grained-intervention}), we created an additional salient object in the middleground of the scene by inserting the object's depth map with increased depth value in the LDM's depth representation. 

\begin{figure}[t]
\begin{center}
\includegraphics[width=0.9\columnwidth]{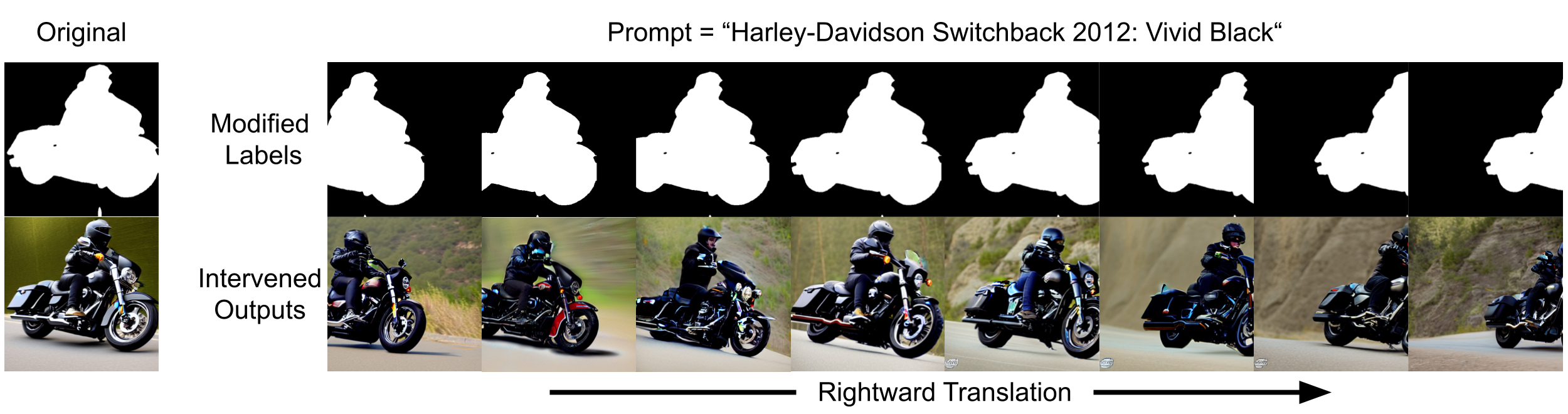}
\caption{A sequence of a moving motorbike created by intervening on the LDM's internal representations of a salient object. All images are generated using the same prompt, initial latent vector, and random seed.}
\label{fig:intervention_application_examples}
\end{center}
\vskip -0.1in
\end{figure}

\section{Related Works}
\label{sec:related-work}
Recently, Stable Diffusion 2.0 introduced the depth as a condition on its text-to-image 2D diffusion model for shape-preserving synthesis. Its depth-conditional model was fine-tuned with a relative depth map as an additional input to the LDM. In a similar vein, Zhao et al.~\cite{zhao2023geofill} suggests using depth estimation as a prior step for better image inpainting. Our work, however, shows that the 2D Stable Diffusion model already encodes the depth dimension even when trained without an explicit depth prior. 

The work of Kim et al.~\cite{kim2022dag}, performed contemporaneously and independently from this work, touches on some of the same issues but from a different perspective -- generating geometrically plausible images. They found a nonlinear depth representation inside a pretrained 2D DDPM using Multi-Layer Perceptrons. Their results also indicated the depth dimension emerges in the early denoising. Our probing experiments, however, suggest that 2D LDM has an even simpler linear depth representation. 

Kim et al.~\cite{kim2022dag} approximated the pseudo ground-truth of depth using predictions from the aggregated representations of multiple feature blocks. They compared the pseudo depth label against a weaker prediction inferred from a single feature block. This difference is applied as a guidance to improve the geometric consistency of output images. From an interpretability perspective, we seek evidence for a causal relation between internal representations and depth. We demonstrate the geometry of images can be altered by directly modifying the LDM's layer-wise internal model of depth, suggesting that a causal role exists. Moreover, our intervention on depth representations can \textbf{control} the scene geometry of output images with respect to a predefined label (see Appendix~\ref{appendix:fine-grained-intervention} and Figure~\ref{fig:intervention_application_examples}). This is not achievable using the guidance methods in~\cite{kim2022dag}. 

Baranchuk et al.~\cite{baranchuk2021label} extrapolated the intermediate activations of a pretrained diffusion model for semantic segmentation. Their high segmentation performance reveals that the diffusion model encodes the rich semantic representations during training for generative tasks. Our work shows that the internal representation of LDM also captures the geometric properties of its synthesized images.

Poole et al.~\cite{poole2022dreamfusion} and Wang et al.~\cite{wang2022score} utilized features learned by 2D diffusion models to optimize a Neural Radiance Field network for 3D synthesis. In contrast, our study centered on finding and interpreting the 3D representations inside LDM. Instead of extending 2D models to a 3D generative task, we take a direct approach of using linear probing classifier to uncover the depth features learned by the self-attention modules.

\section{Conclusion}
\label{conclusion}
Our experiments provide evidence that the Stable Diffusion model, although trained solely on two-dimensional images, contains an internal linear representation related to scene geometry. Probing uncovers a salient object / background distinction as well as information related to relative depth. 
These representations emerge in the early denoising stage. Furthermore, interventional experiments support a causal link between the internal representation and the final image produced by the model. These results add nuance to ongoing debates about whether generative models can learn more than just ``surface'' statistics.

Our experiments also suggest a number of avenues for future research. A natural extension is to look for representations of other scene attributes, such as lighting or texture. Indeed, just as certain language models are said to ``rediscover the NLP pipeline''~\cite{tenney2019bert}, perhaps LDMs recapitulate the standard steps in computer graphics. More generally, one might look for models of semantic aspects of a scene, such as sentiment. 

\section{Acknowledgements}
We would like to thank Catherine Yeh, Shivam Raval, and Aoyu Wu for reading and sharing their feedback on this paper. We also wish to thank Kenneth Li and David Bau who contributed their thoughts to an early draft of this work.

\bibliography{ldm_depth}
\bibliographystyle{plain}

\newpage
\appendix
\onecolumn
\section{Weaker Depth Representations in Convolutional Layers}
\label{appendix:depth-in-conv-vs-sa}
\begin{figure}[h]
\begin{center}
\centerline{\includegraphics[width=\columnwidth]{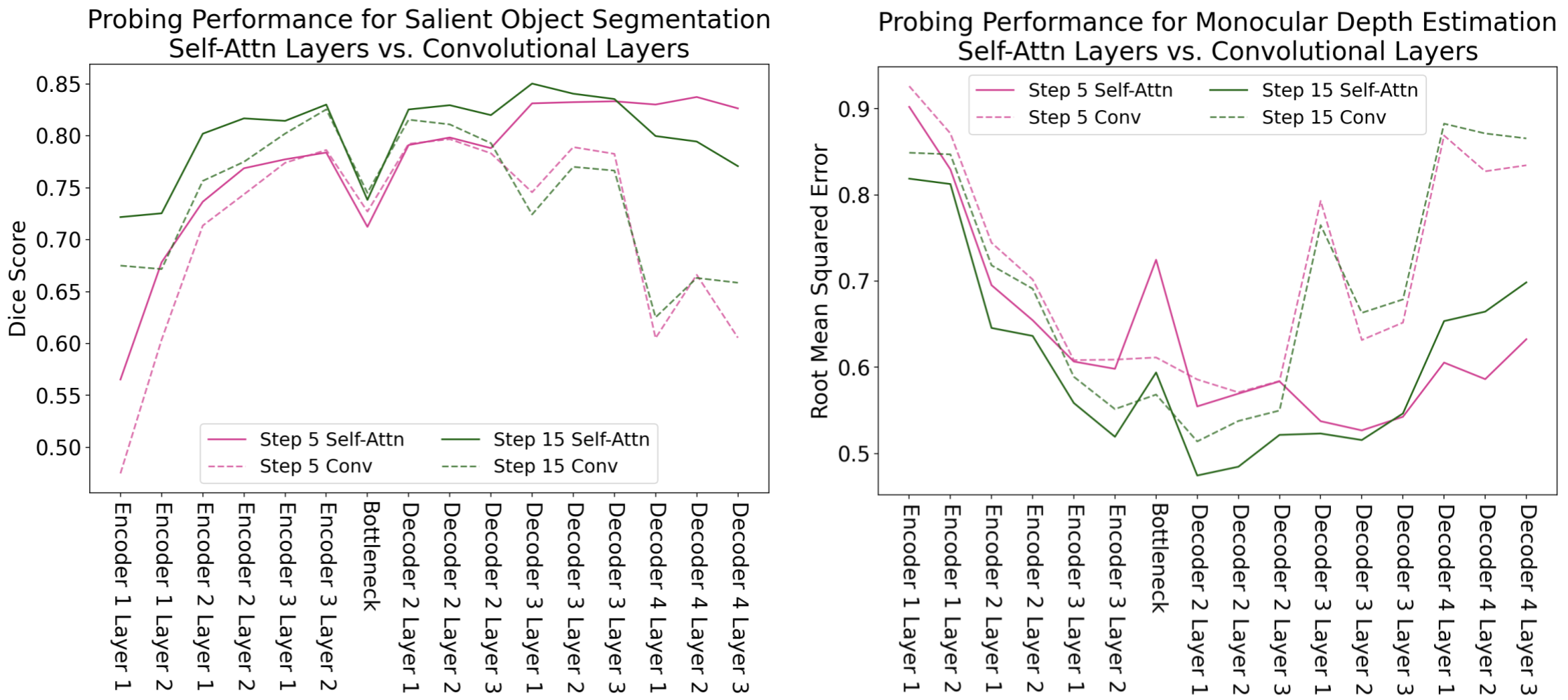}}
\caption{We found stronger depth and salient object representations in the activations of self-attention layers (solid lines). The probing performance on the activations of convolutional layers (dashed lines) is generally weaker.
}
\label{fig:conv-vs-self-attn}
\end{center}
\vskip -0.15in
\end{figure}

In addition to probing the self-attention layers, we also investigated the depth representations in convolutional layers, which produced weaker results. The convolutional layers' performance only starts to improve when the input has been aggressively downsampled (8 x 8 spatial size) such that a 3 x 3 filter kernel can cover a large portion of the input latent image. Intuitively, encoding depth information will require accessing the global context of images, which is only possible when using self-attention.

Our findings revealed that, during generative training without any prior depth information, self-attention layers outperformed convolutional layers in capturing saliency and depth representations.

\section{Spatial and Feature Dimensions of Self-Attention Layers in the LDM}
\label{appendix:spatial-feature-dimension-of-self-attention-layers}

\vskip -0.1in
\begin{table}[h]
\caption{Spatial and feature dimensions of self-attention layer activations across transformer blocks of the LDM.}
\vskip 0.1in
\begin{center}
\begin{small}
\begin{tabular}{lccc}
\toprule
Blocks     & \begin{tabular}[c]{@{}c@{}}Number of \\ Self-Attn Layers\end{tabular} & \begin{tabular}[c]{@{}c@{}} Spatial \\  $h \times w$ \end{tabular}&\begin{tabular}[c]{@{}c@{}} Feature \\ $c$\end{tabular}    \\ \midrule
Encoder 1  & 2                                                                     & 64 $\times$ 64 & 320  \\
Encoder 2  & 2                                                                     & 32 $\times$ 32 & 640  \\
Encoder 3  & 2                                                                     & 16 $\times$ 16 & 1280 \\
Encoder 4  & 0                                                                     & --                & --     \\
Bottleneck & 1                                                                     & 8 $\times$ 8   & 1280 \\
Decoder 1  & 0                                                                     & --               & --     \\
Decoder 2  & 3                                                                     & 16 $\times$ 16 & 1280 \\
Decoder 3  & 3                                                                     & 32 $\times$ 32 & 640  \\
Decoder 4  & 3                                                                     & 64 $\times$ 64 & 320  \\ \bottomrule
\end{tabular}
\end{small}
\end{center}
\label{table:spatial-feature-dimension-of-LDM}
\end{table}

In this section, we review the architecture of Stable Diffusion, which helps explain why we need to upsample the predictions from probing classifier. We will use the information in Table~\ref{table:spatial-feature-dimension-of-LDM} again in Appendix~\ref{appendix:smoothness-regularization}.

As Table~\ref{table:spatial-feature-dimension-of-LDM} shows, the self-attention layers of LDM operate at different spatial and feature dimensions across vision transformer blocks. The probing classifier takes the original activations from the self-attention layers as input, and the classifier outputs the prediction in the same resolution of activations. We upsampled the low resolution predictions to the same spatial size as the original images ($512 \times 512$) when comparing the predictions with the synthetic labels.

\section{Visualizations of Emerging Depth Representations in the LDM}
\label{appendix:vis-of-emerging-depth}

\begin{figure}[h]
\begin{center}
\includegraphics[width=0.55\columnwidth]{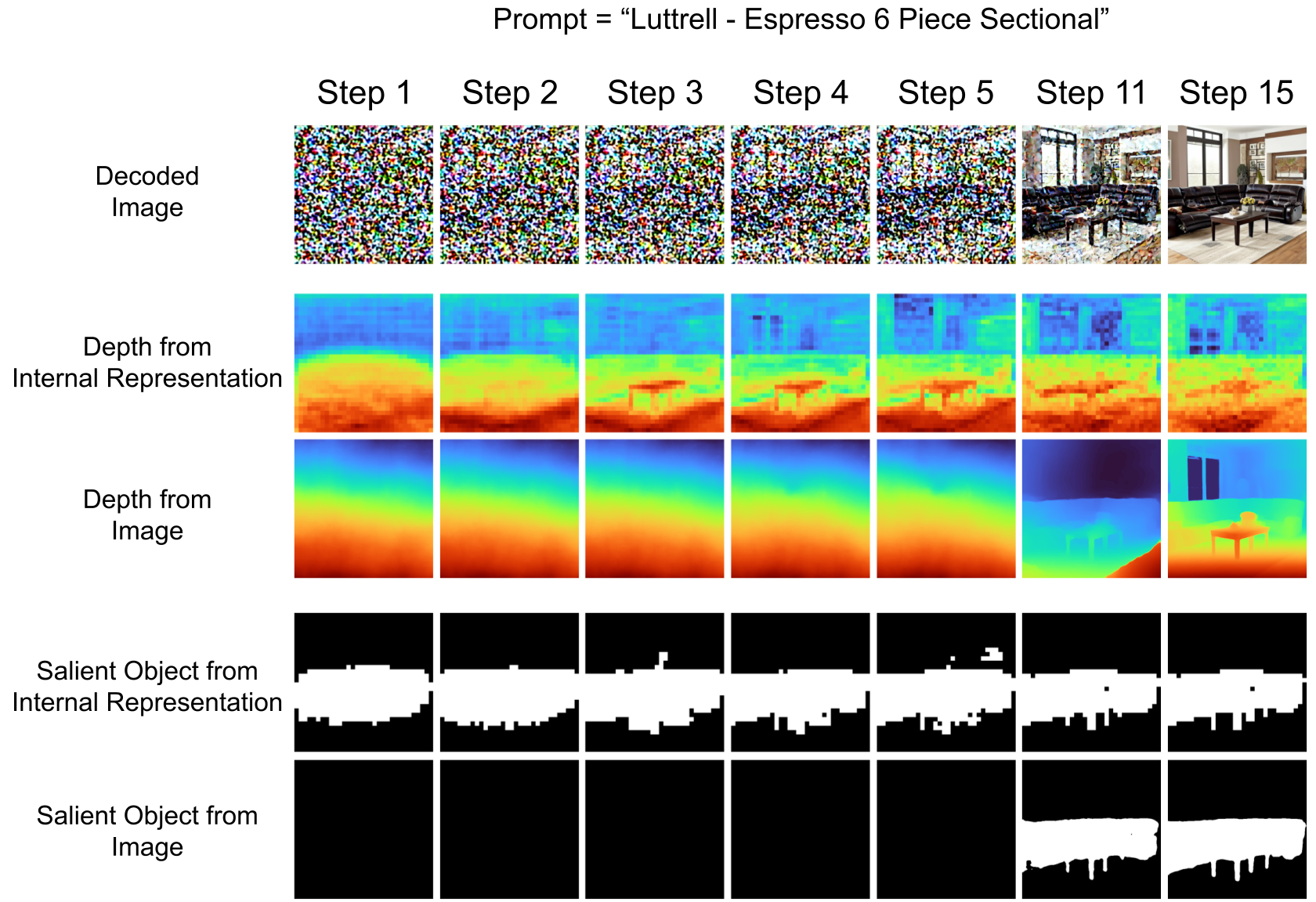}
\caption{Row 1: Images decoded at each denoising step. The latent vector from which the image is decoded serves as the input to the LDM at next step. Row 2 \& 4: Predictions of probing classifiers based on the LDM's internal activations. Row 3 \& 5: Baseline prediction of depth and salient object from external models that take decoded images as input.}
\label{fig:more-examples-of-emerging-depth}
\end{center}
\end{figure}

We observed that the position of the salient object and depth of the scene were both determined at the very early denoising stage (see Figure~\ref{fig:more-examples-of-emerging-depth}). Despite the LDM being conditioned on noisy latent vectors, the probing classifier's predictions of the salient object and depth already resembled those in the fully denoised image.

Visit this \textcolor{blue}{\href{https://drive.google.com/drive/folders/1o0iszSlZcPugvp6mhekcOCEoyDfzdVdo?usp=sharing}{link}} to see more examples of how depth representations develop in the early denoising stages of LDM.

\section{Choices of Intervened Denoising Steps}
\label{appendix:choice-of-intervened-denoising-steps}

\begin{figure}[h]
\begin{center}
\centerline{\includegraphics[width=0.99\columnwidth]{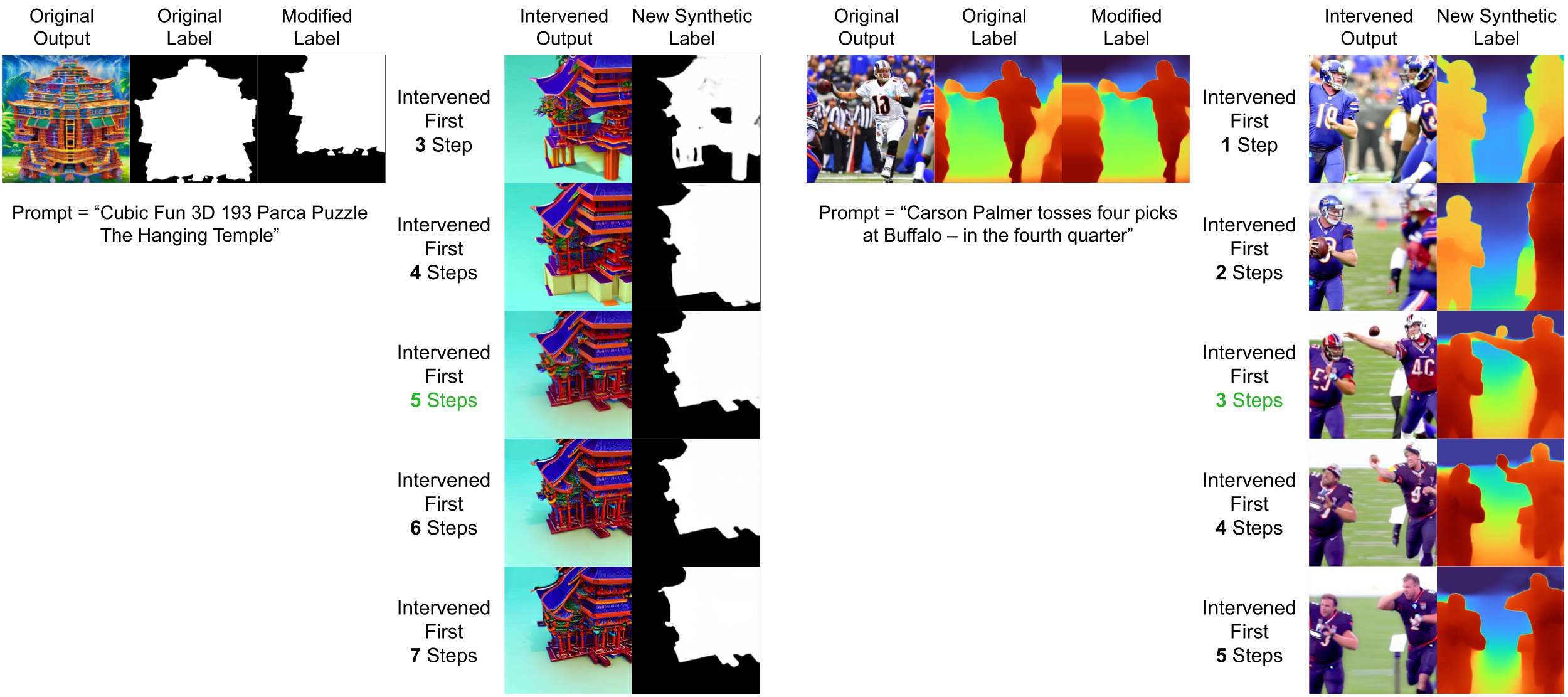}}
\caption{Denoising steps: we experimented with intervening in multiple numbers of steps, both for saliency (left) and depth representations (right). For the saliency representation, the intervention was effective after intervening 5 steps. Further increasing the number of intervened steps had almost no change on the model's output. For the depth representation, the intervention was effective after intervening three steps. Intervening additional steps did not improve the results.
}
\label{fig:choice_of_intervened_steps}
\end{center}
\vskip -0.1in
\end{figure}

One variable in our intervention experiment is how many denoising steps to intervene on. In a preliminary study, we compared the model's outputs after intervening different numbers of denoising steps on saliency and depth representations. 

As Figure~\ref{fig:choice_of_intervened_steps} shows, the intervention on salient object representations was effective after intervening the first 5 steps of denoising. Further increasing the number of intervened steps had no significant influence on the generated image. For the intervention on depth, we observed that intervening the first 3 steps of denoising was sufficient to change the depth of generated scenes.

\section{Fine-grained Intervention on Depth Representation}
\label{appendix:fine-grained-intervention}

\begin{figure}[t]
\begin{center}
\includegraphics[width=\columnwidth]{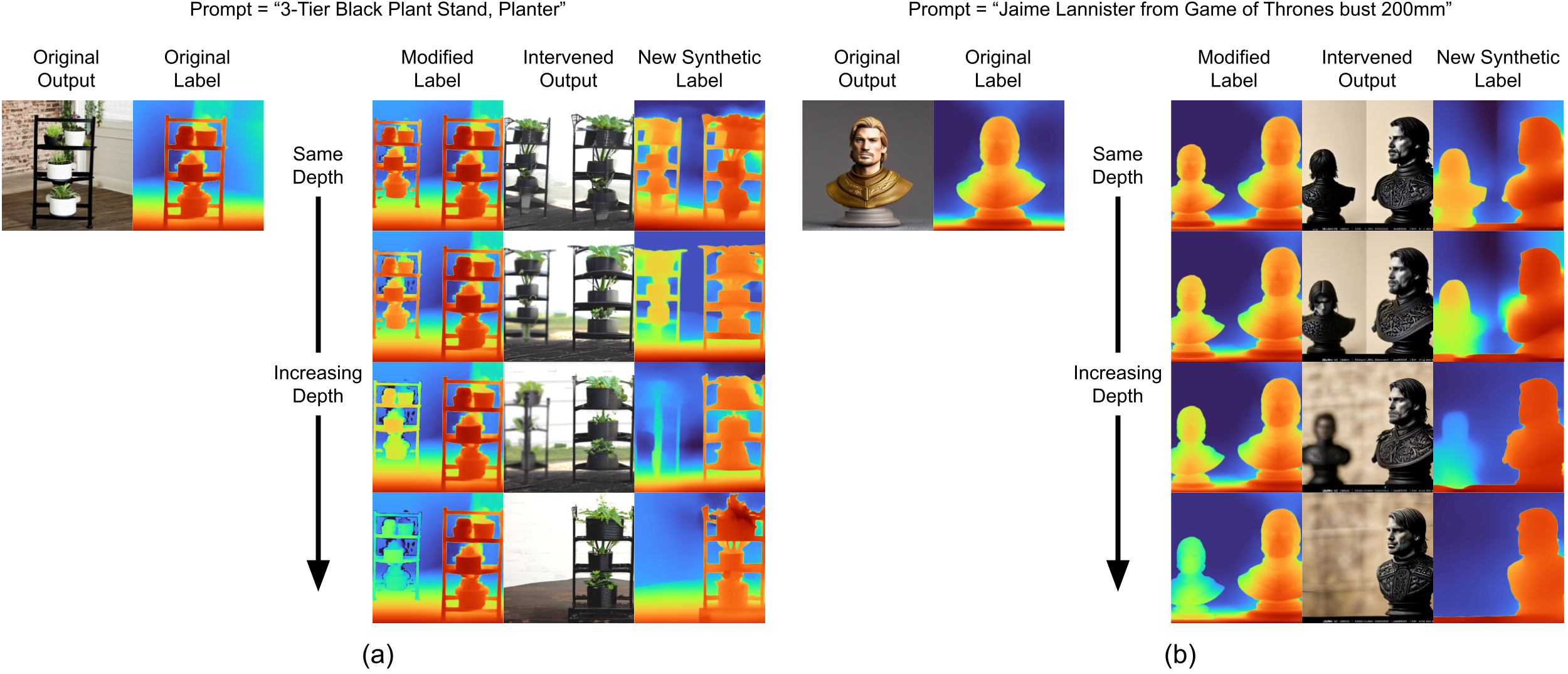}\\
\vskip 0.1in
\includegraphics[width=\columnwidth]{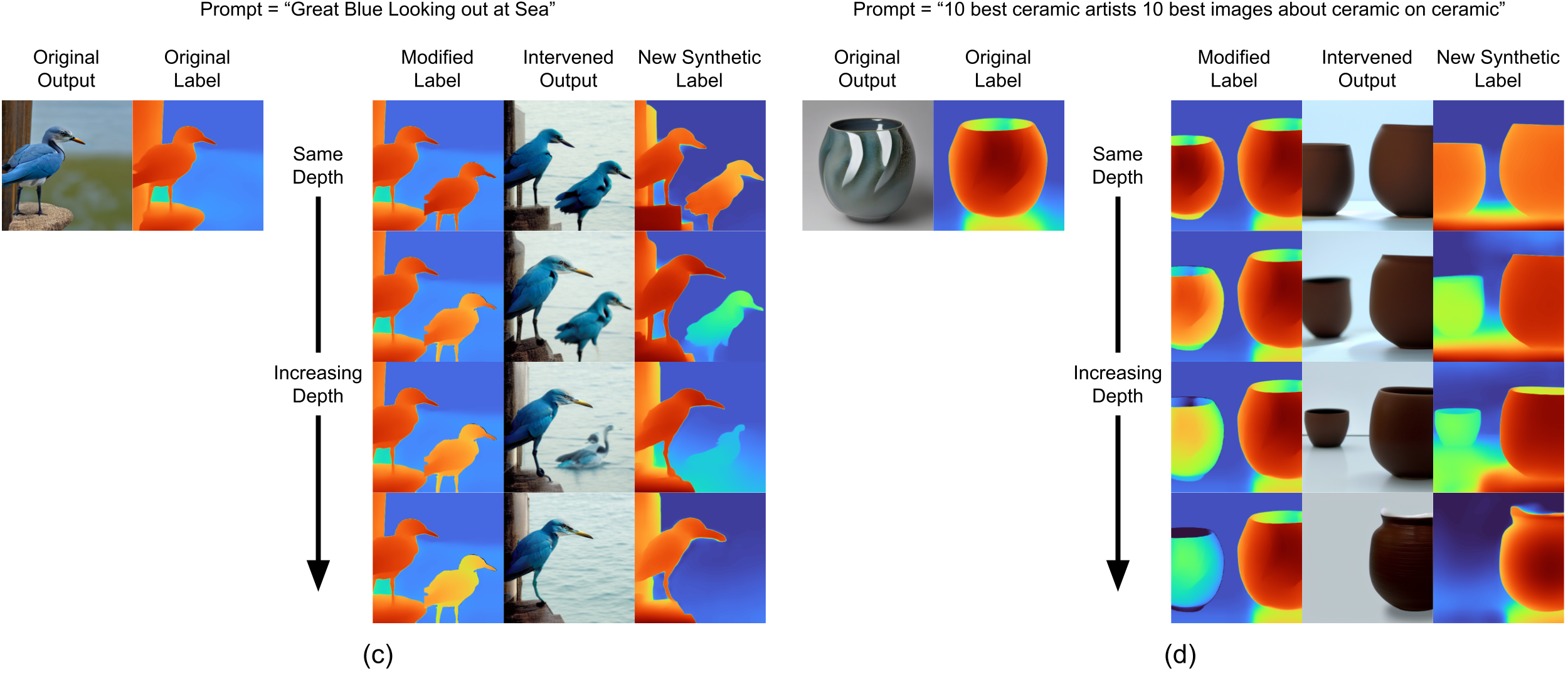}
\caption{To make modified depth labels consistent with the rules of perspective, we scaled down the salient object that was farther away from the viewpoint in the modified label. For each intervention, we progressively increased the depth of the added salient object.}

\label{fig:scaled_down_salient_objects_at_different_depths}
\end{center}
\vskip -0.1in
\end{figure}

\textbf{Adding salient objects at different depths:} In this experiment, we aimed to test the continuity of the LDM's depth representation. Specifically, if we increase the depth disparity between two objects in the LDM's representations, will the depth difference between two objects enlarge accordingly in the output image? 

To perform this intervention, we copied the depth map of a salient object onto a horizontally translated depth label. The modified depth label initially has two salient objects at the same depth. To create the depth disparity, we increased the depth value of the added salient object so it became distant in the depth label. It is arguable that having two objects with the same size at different depths breaks the rules of perspective. To make the modified depth label geometrically coherent, we scaled down the salient object that was distant to the viewpoint. This experiment required manually modifying depth label for intervention, and it is therefore hard to generalize on the entire dataset for quantitative evaluation.

As Figure~\ref{fig:scaled_down_salient_objects_at_different_depths} shows, inserting the depth map of a salient object in the LDM's depth representation created another salient object at the corresponding location in the generated image. Increasing the depth of the added object pushed it farther away from the viewpoint. In Figure~\ref{fig:scaled_down_salient_objects_at_different_depths}ab, increasing the depth of the added objects resulted in a blurred effect and creates the perception of greater depth within the scene. In Figure~\ref{fig:scaled_down_salient_objects_at_different_depths}d, the object with increased depth also exhibited a reduction in its physical size.

\begin{table}[t]
\caption{Smoothness constraint negatively affects the performance of probing regressors, especially when the resolution of probing prediction is low. The resolution of the probing prediction is equal to the spatial size of input activations, which can be as small as $8 \times 8$ (see Table~\ref{table:spatial-feature-dimension-of-LDM} of Appendix~\ref{appendix:spatial-feature-dimension-of-self-attention-layers}). The spatial size of the ground truth label is $512 \times 512$. We observed that training probing regressors without smoothness loss improved the probing performance.}
\vskip 0.1in
\begin{center}
\begin{footnotesize}
\begin{tabular}{@{}ccccccc@{}}
\toprule
\multicolumn{1}{l}{}                                                     & \multicolumn{6}{c}{At Denoising Step 15}                                                                                           \\ \midrule
\multicolumn{1}{l}{}                                                     & \multicolumn{2}{c|}{Encoder Block 3}          & \multicolumn{1}{c|}{Bottleneck} & \multicolumn{3}{c}{Decoder Block 2}              \\
RMSE $\downarrow$                                                        & Layer 1        & \multicolumn{1}{c|}{Layer 2} & \multicolumn{1}{c|}{Layer 1}    & Layer 1        & Layer 2        & Layer 3        \\ \midrule
Without Smoothness                                                       & \textbf{0.559} & \textbf{0.519}               & \textbf{0.594}                  & \textbf{0.474} & \textbf{0.485} & \textbf{0.522} \\
With Smoothness                                                          & 0.583          & 0.545                        & 0.639                           & 0.501          & 0.511          & 0.543          \\
T-test p-value                                                           & < 0.05 & < 0.05               & $\ll$ 0.05                      & $\ll$ 0.05     & $\ll$ 0.05     & < 0.05 \\ \midrule
\begin{tabular}[c]{@{}c@{}}Spatial Size\\ $h \times w$\end{tabular} & 16 $\times$ 16 & 16 $\times$ 16               & 8 $\times$ 8                    & 16 $\times$ 16 & 16 $\times$ 16 & 16 $\times$ 16 \\ \bottomrule
\end{tabular}
\end{footnotesize}
\end{center}
\label{table:remove-smoothness-loss}
\vskip -0.1in
\end{table}

\begin{figure}[t]
\begin{center}
\includegraphics[width=0.8\columnwidth]{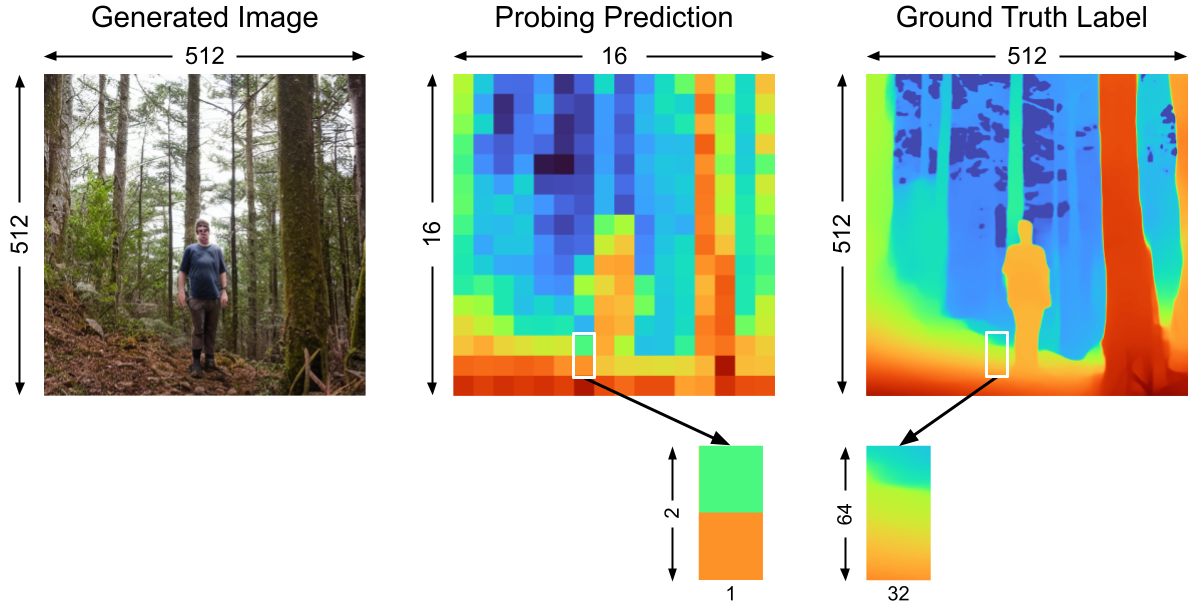}
\caption{The probing predictions have much smaller spatial resolutions compared to that of the generated images. In this example, the probing prediction has a spatial size of only $16 \times 16$, whereas the ground truth depth label of the generated image has a spatial size of $512 \times 512$. Each pixel of the probing prediction represents $32 \times 32$ pixels of the ground truth label. While we anticipate the per-pixel depth to change smoothly in the high resolution depth label, this is not true for the low resolution probing prediction. Applying the smoothness regularization on the low resolution prediction adversely affects probing performance.}

\label{fig:smoothness-constraint-hurt}
\end{center}
\vskip -0.1in
\end{figure}

\section{Smoothness regularization}
\label{appendix:smoothness-regularization}
When training the probing regressors for depth estimation, we also experimented with applying a smoothness constraint~\cite{johnston2020self} to the probing prediction. The local changes in per-pixel depth within a high resolution image are mostly small. The smoothness constraint leverages this property of the image and penalizes rapid local changes to improve the depth prediction. 

However, the depth predictions from our probing regressors have much lower resolution compared to the generated images, since LDM operates on a latent vector with smaller spatial size. In the low resolution probing prediction (see Figure~\ref{fig:smoothness-constraint-hurt}), one pixel in the depth map represents a much larger region in the generated image, and the change in depth between two large regions are often unsmooth. We observed that training the probing regressor without smoothness loss improved its depth estimation performance, especially when the resolution of the probing prediction is low (see Table~\ref{table:remove-smoothness-loss}).

\end{document}